\documentclass[paper=a4]{article}
\pdfoutput=1
\usepackage{PRIMEarxiv}

\usepackage[utf8]{inputenc}
\usepackage[T1]{fontenc}
\usepackage{hyperref}
\usepackage{amsfonts}
\usepackage{fancyhdr}       
\usepackage{graphicx}       
\usepackage{multirow}
\usepackage{float}
\usepackage{caption}
\usepackage{longtable}
\usepackage{amsmath}
\usepackage{dirtytalk}
\usepackage{xcolor}
\usepackage{soul}

\usepackage[
  backend=biber,
  style=ieee,
]{biblatex}
\graphicspath{{.}}

\let\cite=\autocite

\title{Automated Feedback Generation for a Chemistry Database and Abstracting Exercise}

\author{
  Oscar Morris \\
  \texttt{twocap06@gmail.com} \\
  \And
  Russell Morris \\
  EaStCHEM School of Chemistry \\
  University of St Andrews \\
  \texttt{rem1@st-andrews.ac.uk} \\
}

\begin{document}
\maketitle

\begin{abstract}
Timely feedback is an important part of teaching and learning. Here we describe how a readily available neural network transformer (machine-learning) model (BERT) can be used to give feedback on the structure of the response to an abstracting exercise where students are asked to summarise the contents of a published article after finding it from a publication database. The dataset contained 207 submissions from two consecutive years of the course, summarising a total of 21 different papers from the primary literature. The model was pre-trained using an available dataset (approx. 15,000 samples) and then fine-tuned on 80\% of the submitted dataset. This fine tuning was seen to be important. The sentences in the student submissions are characterised into three classes – background, technique and observation – which allows a comparison of how each submission is structured. Comparing the structure of the students’ abstract a large collection of those from the PubMed database shows that students in this exercise concentrate more on the background to the paper and less on the techniques and results than the abstracts to papers themselves. The results allowed feedback for each submitted assignment to be automatically generated.
\end{abstract}

\section{Introduction}

Feedback is recognised as an important part of modern teaching and learning in higher education \cite{evansMakingSenseAssessment2013}. There is no prescriptive definition of what feedback is most appropriate as it can vary between the type of assignment or assessment involved. The best feedback may often be a balance between corrective (cognitivist) and socio-constructivist (i.e. encouraging the student to learn through self- or peer-group evaluation) and may be written or oral \cite{archerStateScienceHealth2010}. However, whatever the method of feedback chosen for any assignment there is no doubt that any feedback should be timely. This is because the best advice for improving student performance, narrowing the gap between real performance and that expected by the instructor \cite{wiliamWhatAssessmentLearning2011} will be when the task is still relatively fresh in their mind. In large modules with many students the workload involved with providing timely feedback can be significant and achieving timeliness can become critical when, for example, there is limited time between an assignment and a final examination. 

Machine learning (ML) is a computer technology that relies on algorithms built by learning from training data.  In recent years there has been a growing interest in using ML as an aid to assessment in STEM education. Zhai and co-workers \cite{zhaiApplyingMachineLearning2020} have recently reviewed approximately 50 papers that used ML as an assessment tool across science, although only one of the papers concerned a specific chemistry content. More recently Yik \textit{et al.} \cite{yikDevelopmentMachineLearningbased2021} have given a compelling argument that ML approaches are of sufficient accuracy (~90\%) in assigning scores to open-ended (formative) assessments (of how students understand acid-base mechanisms) to make them a valuable tool.  However, there remain some ethical issues with assigning marks using such a system (especially in so-called high stakes assignments where the result can affect the degree classification or progression of a student). 

Rather than use ML for automatic scoring \textit{per se}, the aim of the project reported here was to examine whether ML approaches can be used to provide quick, useful feedback for a second-year chemistry assignment, where the feedback could be used by both the student and the human marker assigning the scores for the work. The chosen assignment is one whose aim is to introduce students to databases and to reading primary literature. For most of the class this is the first time they will have encountered such an assignment. However, this is a skill they will use throughout their degree and, in particular, during research-led teaching and research projects. The assignment consists of two parts: a database search and a paper summary/abstract. In the first part of the assignment the students find a particular paper in a database, locate the journal impact factor, reformat the reference in two different styles and then look up how many citations the paper has received. These types of questions are fairly trivial to give feedback for as they essentially have correct answers (or answers in a narrow band of possibilities). These do not require an ML approach to generate feedback. In the second part the students are required to summarise the content of the paper. Free-form questions, such as the summaries in this assignment, are more appropriate for a machine-learning approach \cite{yikDevelopmentMachineLearningbased2021}. Modern methods to approach this problem use neural architectures to automatically learn which features are important \cite{taghipourNeuralApproachAutomated2016}. These systems have been improved by the introduction of the attention mechanism \cite{bahdanauNeuralMachineTranslation2016} which allow a model to assign a weight to the importance of each word in the input text, allowing the model to better extract the meaning of the text. However, an extra challenge in the assessment of this assignment is that students are allocated a different paper to summarise – the project has therefore focused on giving feedback on the structure of the responses.
Automatic systems of this kind could be very useful because they significantly lessen the workload on academic staff and increase the efficiency of students’ learning by providing quick feedback on their work. This could allow students to improve their work without direct feedback from a member of staff, increasing students’ independence and the efficiency of their revision. In addition, it can also be used to flag up quickly those students who need urgent intervention from a member of staff early in the process. For some assignments students may be supported by a subsequent small group tutorial where a final mark is given by the tutor based on both their written response to the assignment and their oral contribution during the tutorial itself. Timely automatic feedback will also support this process as it gives the tutor a guide as to which areas would be most fruitful to discuss in the tutorial.

In the following study we have used a two-step approach to providing feedback. As a check to ensure that the machine learning can indeed identify answers of different quality we used an Automated Essay Scoring (AES) system to provide a mark for the assignment, which was then compared to that provided by the human markers. AES systems have been well researched \cite{alikaniotisAutomaticTextScoring2016, utoReviewDeepneuralAutomated2021, rameshAutomatedEssayScoring2022} and are capable of grading essays almost within the level of disagreement between human markers. However, because of the ethical issues surrounding the implementation of AES we have only used this as a validity check at this time and so 

The second step in the process is Automated Feedback Generation (AFG). In contrast to automatic scoring (AES), AFG has not been studied anywhere near so often \cite{liuAutomatedEssayFeedback2017, tashuSemanticBasedFeedbackRecommendation2020} To provide feedback on the abstracts, we propose a method that uses features extracted by a neural model that are used to provide guiding questions to the student to improve their work and to provide topics for discussion in a subsequent class. Since AES and AFG are very different tasks that require different approaches, we discuss each task in turn and finally provide an example of the feedback on the structure of assignment answer that a student receives using this system.

\section{The Assignment}

The dataset used for these experiments was derived from a chemistry database/abstracting exercise where each tutorial group was given a different paper for which they had to write a summary in abstract form and provide other information. The information the students had to provide is as follows:

\begin{enumerate}
\item The Impact Factor of the Journal
\item The Reference in the Royal Society of Chemistry format 
\item The Reference in the American Chemical Society format
\item The number of times the paper was cited.
\end{enumerate}

Each of these pieces of information are given 0.5 or 1 mark depending on whether they are “partially correct” or “fully correct.” The correct answers for the 4 questions listed above were extracted by selecting the most common answer in the dataset (if this system were deployed, the assessors could choose their own correct answers) as the majority of students get these questions correct.
After finding these 4 pieces of information the students write an abstract/summary for the paper. The prompt they were given is:

\begin{quote}Summarise, in abstract form, the content of the paper.\end{quote}

The abstract is awarded 0-6 marks providing a total of 10 marks for the exercise.

The task can then be further broken down such that the AES system first provides a mark and feedback on the accuracy of the 4 pieces of information and then provides a mark and feedback on the abstract.

\section{The Dataset}

The dataset contains 207 samples sourced from two consecutive years’ answers to the assignment. The task involved 21 different papers from the primary literature so each tutorial group has a different one. To evaluate the model the dataset was split into two parts, 80\% of the data was used to train the neural model and the other 20\% was used to evaluate the AES system. Evaluation of the automatic feedback was performed manually as there was no human-given feedback with the samples.

\section{The Automatic Scoring System}

In the same way that the exercise contains two distinct types of question, these questions are scored individually by two different systems and then the scores are added to produce the final mark out of 10.

\subsection{Scoring Parts 1 to 4}

The numerical questions are, in principle, very simple to score. The current impact factor question (part 1) should have a definitively correct answer while the number of times cited (part 4) response is time sensitive: depending on the exact date that the student looked up the information the database may have been updated. One could give either 0 or 1 depending on certain criteria. However, to facilitate the feedback aspects we chose to give three possible marks. The value given by the student is compared to the correct value and if the student’s value is within 10\% percentage difference of the correct value the answer is deemed “fully correct” and one mark is given. If the student’s value is between 10\% and 25\% percentage difference of the correct value then the answer is deemed “partially correct” and 0.5 marks are given. If the student’s value is outside 25\% then the answer is deemed “incorrect” and no marks are awarded. 

For the referencing formatting questions (parts 2 and 3) a different approach is required. One method to mark such textual questions is to extract the information from the reference e.g. Authors, Year, Journal etc. check that they are correct and in the correct order. However, perhaps a simpler method to implement is to use the cosine similarity, as defined in \cite{tashuSemanticBasedFeedbackRecommendation2020}, to compare the two texts. A cosine similarity higher than 0.9 is deemed fully correct and one mark assigned, a cosine similarity between 0.65 and 0.9 is deemed partially correct earning a mark of 0.5 and a cosine similarity less than 0.65 is deemed incorrect and gaining zero marks.

\subsection{Scoring the Abstracting Assignment: the Pre-Training Dataset}

Because of the small dataset of only ~200 samples, we decided to pre-train the model on a much larger dataset. For this the datasets created for the Automated Student Assessment Prize (ASAP) sponsored by the Hewlett Foundation were used. For these competitions two datasets were published, one dataset for Automated Essay Scoring \cite{hamnerHewlettFoundationAutomated2012} and one for Short Answer Scoring (SAS) \cite{hamnerHewlettFoundationShort2012}. These datasets contain approximately 15,000 samples each. This allows for the AES model to begin with a much better ‘foundation’ when it begins to score abstracts and means that the model can achieve high performance while not overfitting to the training set. Overfitting is a particular problem as it means that the model cannot generalise from the training set. In effect, overtraining means the system is ‘memorising’ each sample in the training set rather than finding patterns.

The pre-training dataset contain texts from a variety of subjects including Physics, English, Biology and Chemistry among others. This variety makes a combined dataset suitable for training a subject-agnostic AES model that can be fine-tuned on a specific question with a small dataset to achieve higher performance. Each sample was min-max normalised to a range [0,1]; this was done according to the question the sample was answering, effectively converting each mark to a percentage, i.e. a fully correct answer achieves a score of 1, regardless of how many total marks are available for the question.

\subsection{The AES Model}

The model used for the AES portion of the task uses a Bidirectional Long Short-Term Memory (BiLSTM) \cite{hochreiterLongShortTermMemory1997} encoder with an attention mechanism \cite{bahdanauNeuralMachineTranslation2016}. The input sequence is tokenised using the same tokeniser as used in the BERT transformer model \cite{devlinBERTPreTrainingDeep2019}. This tokeniser is used because of its ability to tokenise to word parts rather than whole words. This means that when technical language is used, as is often the case with texts such as abstracts, no vocabulary needs to be added, this then improves the effectiveness of pre-training such a model.

\subsection{Metrics}

\subsubsection{Training Metric}

To train the AES model a dynamic loss function was used that combines the error in the standard deviation of the model’s predictions and the Mean Squared Error (MSE) metric \cite{morrisEffectivenessDynamicLoss2023}. Multiple loss functions can be combined using a weighted sum. This weight, p is defined in Eq.\ref{eq:min-decay}.

\begin{equation} p(t) = \min\left(a, a \cdot \exp\left(-c \cdot \left(\frac{t}{T} - b \right)\right)\right) \label{eq:min-decay} \end{equation}

where \(a\), \(b\) and \(c\) are constants, \(t\) is the current training step and \(T\) is the total number of training steps. Such that:

\begin{equation} L_T(t) = p(t) \cdot L_1 + (1-p(t))\cdot L_2 \label{eq:combine-loss} \end{equation}
 
where \(L_T\) is the total loss, and \(L_1\) and \(L_2\) are two loss functions. For this dynamic loss function \(L_1\) is the error in the standard deviation as defined in Eq. \ref{eq:std-error} and \(L_2\) is the MSE metric as defined in Eq. \ref{eq:mse}:

\begin{equation} \text{STDE}(\mathbf{x}, \mathbf{y}) = |\sigma(\mathbf{y}) - \sigma(\mathbf{x}) \label{eq:std-error} \end{equation}

where \(mathbf{y}\) are the model's predictions, \(mathbf{x}\) are the correct values and \(\sigma\) is the function calculating the standard deviation.

\begin{equation} \text{MSE}(\mathbf{x}, \mathbf{y}) = \frac{1}{N} \sum^{N}_{i=0}(y_i - x_i)^2 \label{eq:mse} \end{equation}

\subsubsection{Evaluation Metrics}

To evaluate the performance of the AES model several metrics are used, primarily the coefficient of determination \(r^2\) as defined in Eq. \ref{eq:r2} and also the Mean Absolute Error (MAE), defined in Eq. \ref{eq:mae} and Root Mean Squared Error (RMSE) defined in Eq. \ref{eq:rmse}.

\begin{equation} r^2 = 1- \frac{\sum^{N}_{i=0}(y_i - x_i)^2}{\sum^{N}_{i=0}(y_i-\bar{y})^2} \label{eq:r2} \end{equation}

where \(\bar{y}\) is the mean of the predicted values.

\begin{equation} \text{MAE}(\mathbf{x}, \mathbf{y}) = \frac{1}{N} \sum^{N}_{i=0}|y_i-x_i| \label{eq:mae} \end{equation}

\begin{equation} \text{RMSE}(\mathbf{x}, \mathbf{y}) = \sqrt{\frac{1}{N} \sum^{N}_{i=0}(y_i - x_i)^2} \label{eq:rmse} \end{equation}

\subsection{Results}

After fine-tuning the neural model, the entire AES system was tested on the evaluation split of the dataset. The model was also evaluated after only being pre-trained on the combined ASAP dataset. The model was unable to learn only using the small fine-tuning dataset due to severe overfitting. Table \ref{tbl:aes-results} shows the performance of both models at each stage of training. 

\begin{table}[H]
\centering
\captionsetup{justification=centering}
\caption{Performance of the model after only Pre-Training and after Pre-Training followed by Fine-Tuning. These values are the mean of 5 runs. \label{tbl:aes-results}}
\begin{tabular}{|l|cccc|}
\hline
Model       & $r^2$ & MAE  & RMSE & Maximum Error \\ \hline
Pre-Trained & 0.29  & 0.12 & 0.15 & 0.28          \\
Fine-Tuned  & 0.51  & 0.10 & 0.12 & 0.23          \\ \hline
\end{tabular}
\end{table}

These results show that fine-tuning a pre-trained model, even though it has been pre-trained on a very similar task, significantly increases the final performance of the model.

\begin{figure}[H]
\captionsetup{justification=centering}
\centering
\includegraphics[height=6cm]{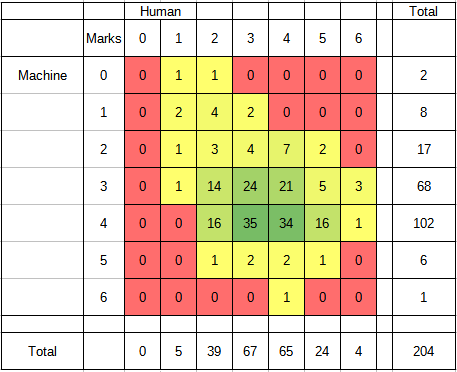}
\caption{Confusion matrix for automatic (machine) scoring system for the abstract section of the assignment compared with the scoring by human markers.}
\label{fig:confusion}
\end{figure}

Figure \ref{fig:confusion} shows a confusion matrix for this run of the model. It shows the correlation between the results obtained from human marking and the automatic scoring. A perfect correlation between the two would give only scores on the diagonal of the matrix. Here we see that the automatic scoring gives considerably more marks of 4 out of 6 than do the human markers and also more 2 out of 6. The spread of marks for the human marking seems a little greater than for the automatic machine marking. However, given that for ethical reasons we do not recommend using machine-generated marking at the present time, these results were deemed to be promising enough to move forward to see if the desired automatic feedback could be machine generated. It should also be noted that the marking has not been optimised fully for scoring and so the correlation may well be improved by further work. For the overall goal of this project this was not deemed necessary.

\section{The Automatic Feedback Generation System}

As described above, the assignment is split into two quite different tasks and the AFG system can similarly be split into two components. To provide simple cognitivist feedback on the first four questions is mostly trivial as it is based on whether the answers are right or wrong. The feedback can then be chosen from a pre-prepared list of comments. Comments such as “Make sure your Royal Society of Chemistry reference has exactly the correct format” or “That is the correct Impact Factor, Well done!” can be given simply on the basis of whether the answers given are right or wrong. This form of feedback (rather than simply saying whether the answer is right or wrong) suggests to the student that they should review their work more closely. 

For more free form answers to the abstracting/summarising section the feedback provided must be more nuanced. This is particularly true in an assignment where the students are not all summarising the same paper – each group has a different topic. However, the structure of a response is likely to be similar whatever the scientific content of the paper. Therefore, to provide feedback on the structure of the abstracts we decided to classify each sentence in the abstract and provide feedback based on how each sentence was classified. To provide feedback on the abstracts submitted in this assignment we trained a neural model to classify each sentence in an abstract into one of these categories and then used the results as a guide as to the feedback that should be given. To achieve this level of classification a BERT encoder was used with a classification head designed for multilable classification \cite{devlinBERTPreTrainingDeep2019} as exposed by the simpletransformers python module \cite{rajapakseSimpleTransformers2021}.

\subsection{Training}

To fine-tune this BERT model to classify abstract sentences a dataset was created based on the PubMed20kRCT (a smaller variant of the PubMed200k-RCT dataset)  containing 20,000 abstracts from manuscripts on the subject of medical science that are in the PubMed database \cite{dernoncourtPubMed200kRCT2017}. Despite these abstracts being primarily from the medical field the assumption used in deciding to use this training set is that the structure of an abstract of a scientific paper is relatively invariant irrespective of the specific scientific field. 

The Pubmed20k-RCT dataset classifies abstract sentences into 5 classes: background, objective, method, result and conclusion. Banerjee \textit{et. al} \cite{banerjeeSegmentingScientificAbstracts2020} defined an alternative and slightly simpler approach with only three categories of abstract sentences: background, techniques and observations (results). Table \ref{tbl:mapping} shows how the classifications used in the Pubmed20k-RCT work map onto those used by Banerjee. Our starting point for feedback is that a ‘good’ abstract should have a structure that includes sentences from all three of the Banerjee classes. An abstract that is missing one class of sentence probably does not reflect the content of the scientific paper in enough detail to make it a useful abstract.

\begin{table}[H]
\centering
\captionsetup{justification=centering}
\caption{Mapping between PubMed categories (left) and categories used by Banerjee \textit{et. al} (right) \label{tbl:mapping}}
\begin{tabular}{|l|l|}
\hline
BACKGROUND & \multirow{2}{*}{BACKGROUND}  \\ \cline{1-1}
OBJECTIVE  &                              \\ \hline
METHOD     & TECHNIQUE                    \\ \hline
RESULT     & \multirow{2}{*}{OBSERVATION} \\ \cline{1-1}
CONCLUSION &                              \\ \hline
\end{tabular}
\end{table}

The BERT model was fine-tuned trained on the PubMed dataset for 5 epochs with a batch size of 64. For the purposes of evaluation during training the dataset was split into a training split and an evaluation split, 90\% of the data was used for training the model and the other 10\% was used to evaluate the model.

\subsection{results}

On the evaluation split of the PubMed dataset the model achieved an accuracy score of 92.0\%, meaning that it classified all but 8\% of sentences ‘correctly’. This is very close to the 92.1\% accuracy reported by Banjeree et. al on their PubMed20k-non-RCT. 

When this model is applied to our dataset the model classifies most sentences as either Background or Observation, only 11.5\% of sentences are classified as Technique and 71.5\% of the abstracts only contain sentences in 2 of the 3 classes.  Table \ref{tbl:abstr-dist} shows the distribution of predicted labels throughout the dataset and how students in the assignment strongly emphasise background in their submissions. This is an interesting result in itself in that it shows that when summarising the work the students concentrate on background (and motivation) more than authors do when they write abstracts of their own papers. This is not a great surprise, but of course does reinforce the need for proper setting of expectations for the students before they complete the assignment to make sure they understand what a good structure looks like. It may be perfectly appropriate for the student to write more of a summary rather than an abstract, but the wording of the question, and the instructions/training given to the students needs to be clear as to what the assignment is testing and the types of response expected.

\begin{table}[H]
\centering
\captionsetup{justification=centering}
\caption{Distribution of predicted labels on student-written abstracts from our data and abstracts from PubMed20k-RCT \label{tbl:abstr-dist}}
\begin{tabular}{|l|cc|}
\hline
Label       & Our Data & PubMed20k-RCT  \\ \hline
BACKGROUND  & 49.9     & 19.8           \\ \hline
TECHNIQUE   & 11.5     & 33.0           \\ \hline
OBSERVATION & 38.6     & 47.3           \\ \hline
\end{tabular}
\end{table}

Below is an example abstract containing at least one sentence from each category with sentences predicted as BACKGROUND being highlighted \sethlcolor{yellow}\hl{yellow}, sentences predicted as TECHNIQUE being highlighted \sethlcolor{green}\hl{green} and sentences predicted as OBSERVATION being highlighted \sethlcolor{pink}\hl{pink}:

\begin{quote}
\sethlcolor{yellow}\hl{Nitro-anilides are compounds essential for manufacturing a variety of chemical compounds, including pharmaceuticals, dyes and explosives.} \sethlcolor{yellow}\hl{Traditional methods use concentrated sulfuric and nitric acids, where reaction conditions are harsh and have low tolerance for other functional groups.} \sethlcolor{green}\hl{They were replaced by using nitrate salt reagents, which can be difficult to prepare and, from the resultant metal oxides, have a low atom economy.} \sethlcolor{yellow}\hl{More recent methods use AgNO$_2$ to selectively nitrate arenes, but this method uses high quantities of rare metals, which is unsustainable.} \sethlcolor{pink}\hl{This reaction uses NaNO$_2$ and K$_2$S$_2$O$_8$ in CH$_3$CN, using catalytic AgNO$_2$, to selectively nitrate the ortho positions on a variety of arenes, displaying high regioselectivity and chemoselectivity as well as moderate to high yields with a selection of substitutions on the starting anilide.} \sethlcolor{yellow}\hl{The mechanism proceeds through silver chelation and a subsequent radical mechanism, and the silver catalyst is regenerated by the K$_2$S$_2$O$_8$ oxidant.}
\end{quote}

As seen above, 4 out of the 6 sentences are labelled BACKGROUND with only one sentence each fitting into the other two categories.

The feedback given to this abstract is as follows:

\begin{quote}
Your discussion of the paper’s background has a good amount of detail. \\
It might be useful to outline the Techniques the model uses in a bit more detail. \\
It may be worth making sure that the discussion of the conclusions of the paper are clearer. \\
\end{quote}

\subsection{Example 2}

In this section we provide an example of the whole exercise and the marks and feedback our system generated for it.

\subsubsection{Student Response}

\begin{quote}
Journal "Impact Factor": 6.005 \\
\\
Reference in RSC format: \\
A. Lator, S. Gaillard, A. Poater and J.-L. Renaud, \textit{Organic Letters}, 2018, \textbf{20}, 5985–5990. \\
\\
Reference in ACS format: \\
Lator, A.; Gaillard, S.; Poater, A.; Renaud, J.-L. \textit{Well-Defined Phosphine-Free Iron-Catalyzed N-Ethylation and N-Methylation of Amines with Ethanol and Methanol.} Organic Letters 2018, 20 (19), 5985–5990. \\
\\
Number of times cited: 10 \\
\\
Abstract: \\
To combat the issues of toxic chemicals and by-products, noble and precious metal catalysts, and expensive phosphorus ligands, a new method of alkylation of amines was devised. N-ethylation and N-methylation of a broad range of aliphatic and aromatic compounds were demonstrated using a (cyclopentadienone) iron tricarbonyl complex under basic conditions. These compounds were ethylated or methylated using ethanol or methanol. The use of methanol was more energetically demanding due to its higher enthalpy of dehydrogenation. Consequently, a change in hydrogen pressure was required for selective dehydration over dehydrogenation to methylate some compounds. The method shown produced mono- or dialkylated compounds in high yields. DFT calculations revealed potential pathways for the reaction and highlighted the role of hydrogen pressure in driving the equilibrium towards one intermediate and hence the reduction of imines.
\end{quote}

\subsubsection{Our Evaluation}

\subsubsubsection{Marks}

Impact Factor: 1 mark \\
Reference in RSC format: 1 mark \\
Reference in ACS format: 1 mark \\
Number of times Cited: 0 marks, the correct answer is 42, you gave 10

Abstract: 3 marks

\subsubsubsection{Feedback}

\begin{quote}
\sethlcolor{yellow}\hl{To combat the issues of toxic chemicals and by-products, noble and precious metal catalysts, and expensive phosphorus ligands, a new method of alkylation of amines was devised. N-ethylation and N-methylation of a broad range of aliphatic and aromatic compounds were demonstrated using a (cyclopentadienone) iron tricarbonyl complex under basic conditions. }\sethlcolor{green}\hl{These compounds were ethylated or methylated using ethanol or methanol. }\sethlcolor{pink}\hl{The use of methanol was more energetically demanding due to its higher enthalpy of dehydrogenation. Consequently, a change in hydrogen pressure was required for selective dehydration over dehydrogenation to methylate some compounds. The method shown produced mono- or di- alkylated compounds in high yields. DFT calculations revealed potential pathways for the reaction and highlighted the role of hydrogen pressure in driving the equilibrium towards one intermediate and hence the reduction of imines.} \\
\\
\sethlcolor{yellow}\hl{BACKGROUND} \sethlcolor{green}\hl{TECHNIQUE} \sethlcolor{pink}\hl{OBSERVATION} \\
\\
A more balanced discussion of the background of the paper, the techniques of the paper and the observations and conclusions the paper made might improve your work. \\
It might be worth outlining the methods of the paper in greater detail. \\
The abstract contains discussion of each aspect of the paper in a logical order.
\end{quote}

\section{Discussion}

AES and AFG systems have many advantages: they allow for students to receive immediate results, academic institutions can significantly reduce the “unnecessary” time allocated to students in manually marking their work, governmental organisations responsible for nation-wide exams could cut spending on examiners drastically and reduce the stressful period where students are waiting for results, student’s revision could be streamlined by increasing the number of practice essays they could write, receiving immediate feedback and improving their work before an exam. With all of these advantages it may seem perplexing that these systems have not yet been adopted by major organisations and academic institutions, even when these systems are reaching near human-level performance.

Despite all of these advantages, AES systems in particular do have some disadvantages: most notably their lack of transparency, especially with neural models. This is an issue with almost every neural network, they are effectively a ‘black box.’ It is impossible to tell how the model is coming to its conclusions. It is likely that this lack of transparency is the reason why these systems have not left the research phase yet as this will decrease the confidence that students have in the system’s grades. 

These disadvantages are less obvious for AFG systems as there are fewer confidence issues than there are with grades. This means they are likely to be easier to implement. The question is then how can they be used and how should they be used. In the example we have described in this paper we have chosen to concentrate on the structure of the assignment response and so feedback on the content of the response would still have to be completed by a human marker. However, high quality feedback on the structure would still be a very useful starting point for the markers, saving them time in what can be an onerous task if numbers in the class are high. In the first instance, therefore, we would see this feedback going to the human markers first, to be checked and added to before being forwarded to the student. As well as saving the human marker some time, it will also allow confidence in the quality of the feedback to be developed over time. In the long run the goal is to give the students instant, high-quality feedback on their work without the need for intervention by a human marker. This would be particularly advantageous if first drafts of work are passed through the system, which would then allow the students to improve the quality of their assignments before admission. Eventually, it may even be possible to add in automatic marking using an AES system.

\section{Conclusion}

We have shown how available machine learning algorithms can be utilised to give automatic feedback on a chemistry assignment that mixes different types of responses. The most interesting aspect is that there is scope to give instant feedback not only on the parts of the assignment that have objectively correct answers, but that machine learning can be used to give quick feedback on the structure of a response, even if the chemical content of each response differs.

\printbibliography

\end{document}